\pdfoutput=1

\documentclass[11pt]{article}

\usepackage[]{emnlp2021}

\usepackage{times}
\usepackage{latexsym}

\usepackage[T1]{fontenc}

\usepackage[utf8]{inputenc}

\usepackage{microtype}
\usepackage{graphicx}
\usepackage{mathrsfs}
\usepackage{dutchcal}
\usepackage{amstext}
\usepackage{booktabs}
\usepackage{amsfonts}
\usepackage{amsmath}
\usepackage{multirow}
\usepackage{subfigure}
\usepackage{array}
\usepackage{enumitem}
\title{Exploiting Reasoning Chains for Multi-hop Science Question Answering\thanks{\hspace{2mm} The work described in this paper is substantially supported by a grant from the Research Grant Council of the Hong Kong Special Administrative Region, China (Project Code: 14200719)}}

\author{
Weiwen Xu, Yang Deng, Huihui Zhang, Deng Cai and Wai Lam \\
The Chinese University of Hong Kong \\

{\tt \{wwxu,ydeng,hhzhang,wlam\}@se.cuhk.edu.hk} \\
{\tt thisisjcykcd@gmail.com}
}

\begin{document}
\maketitle
\begin{abstract}
We propose a novel Chain Guided Retriever-reader ({\tt CGR}) framework to model the reasoning chain for multi-hop Science Question Answering. Our framework is capable of performing explainable reasoning without the need of any corpus-specific annotations, such as the ground-truth reasoning chain, or human-annotated entity mentions. 
Specifically, we first generate reasoning chains from a semantic graph constructed by Abstract Meaning Representation of retrieved evidence facts. A \textit{Chain-aware loss}, concerning both local and global chain information, is also designed to enable the generated chains to serve as distant supervision signals for training the retriever, where reinforcement learning is also adopted to maximize the utility of the reasoning chains. 
Our framework allows the retriever to capture step-by-step clues of the entire reasoning process, which is not only shown to be effective on two challenging multi-hop Science QA tasks, namely OpenBookQA and ARC-Challenge, but also favors explainability.
\end{abstract}
\section{Introduction}
\begin{figure}[t]
    \centering
\includegraphics[scale=0.47]{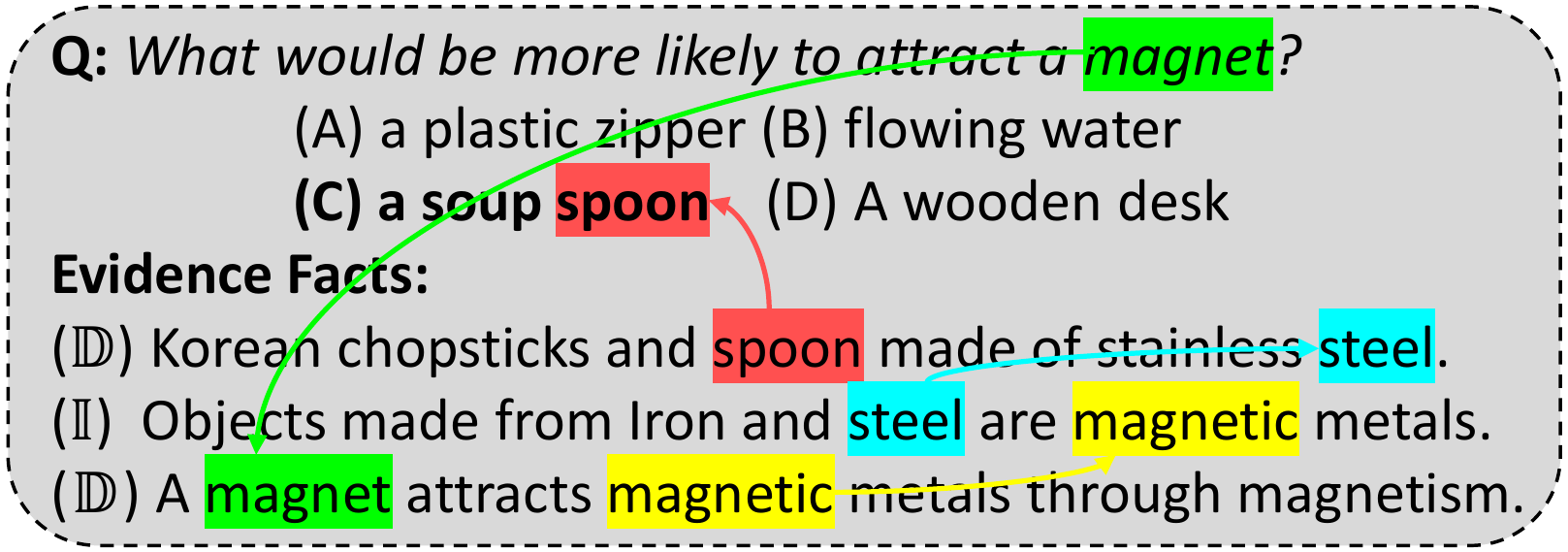} 
    \caption{An example of multi-hop QA with direct ($\mathbb{D}$) and indirect ($\mathbb{I}$) facts to form a reasoning chain.}
    \label{fig:example}
\end{figure}

Question Answering (QA) with external knowledge has gained increasing attention in recent years as it mimics human behavior to first filter out relevant knowledge from massive information. Prior works usually employ a retriever-reader architecture~\cite{chen-etal-2017-reading}, where the retriever retrieves top-ranked evidence facts from a large corpus and the reader conducts reasoning with these facts. This architecture works well in single-hop QA, where the answer can be easily inferred with only one evidence fact. However, it is hard to retrieve all necessary evidence facts to confidently answer a complex question requiring multi-hop reasoning~\cite{shao2021memory}. As shown in Figure~\ref{fig:example}, multi-hop QA usually involves a sequential nature of evidence facts to form a reasoning chain, including (1) direct facts sharing a semantic relationship with the question or the answer; (2) indirect facts sharing little lexical or semantic overlap but serving an irreplaceable role to infer the answer.

A common practice for forming such reasoning chains for multi-hop QA is to expand the chain with iterative retrieval~\cite{xiong2021answering} or sample from an existing or pre-constructed Knowledge Graph (KG)~\cite{Asai2020Learning,yasunaga2021qa}. On one hand, iterative retrieval allows the retriever to capture the evidence-evidence interactions by reformulating the query with newly retrieved evidence fact. However, the retriever would inevitably retrieve partially related facts. Such noise is continuously amplified during the iterative retrieval process, raising concerns about the quality of the reasoning chain. Prior works address this issue by training the retriever against the ground-truth reasoning chain~\cite{yang-etal-2018-hotpotqa,geva2021did}. However, such method is less effective when the ground-truth reasoning chain is partially provided~\cite{OpenBookQA2018,ferguson-etal-2020-iirc} or not applicable when the ground-truth chain is unavailable~\cite{clark2018think}. On the other hand, KG maintains a good growing direction for the reasoning chain. But building a KG usually involves corpus-specific annotations, such as document-level hyperlinks or annotated entity mentions. These limitations make it less applicable in new domains, where hyperlinks and entities are not prevalent~\cite{xiong2021answering}.
 
To address the aforementioned concerns for multi-hop QA, we propose a novel framework, Chain Guided Retriever-reader ({\tt CGR}), to model the reasoning chains, which is compatible without the ground-truth reasoning chain and applicable to broader textual data. In specific, the proposed framework consists of three components: a retriever, a reasoning chain generator, and a reader. The retriever first iteratively retrieves all possible evidence facts. Then the reasoning chain generator adopts Abstract Meaning Representation (AMR)~\cite{banarescu-etal-2013-abstract} to automatically construct a semantic graph that represents the relationship among the retrieved facts. We use AMR because it is not specifically designed for a particular domain and can be adapted to a wide range of sentences  (e.g. Science with little named entity). The final generated reasoning chains are supposed to connect the question nodes and the answer nodes on the semantic graph, such that: (i) the generated reasoning chains serving as a byproduct of our framework can support explainability. (ii) the evidence facts on these chains provide a more appropriate context for the reader because they together fill the knowledge gap between the question and the answer. (iii) the reasoning chains can be used as distant supervision signals to train the retriever in case the ground-truth reasoning chain is not available. 
To achieve these merits, a novel chain-aware loss is proposed to adaptively model the reasoning chains in both supervised and distantly supervised manners. 
The chain-aware loss not only adopts Reinforcement learning (RL)~\cite{Williams-reinforcement} to maximize the utility of the local information from some certain chains based on the reward from the reader, but also enables the retriever to retrieve indirect evidence facts by considering the global information from all the generated chains.
The contributions are summarized as follows:
\begin{itemize}[leftmargin=3mm]
    \item  Our {\tt CGR} framework provides a novel formulation to model the reasoning chains, allowing explainable reasoning without the need of any corpus-specific annotations.
    \item A novel chain-aware loss exploiting both local and global information of reasoning chains is developed to train the retriever, such that the retriever can adapt its retrieval policy to allow high-quality reasoning chains to be generated.
    \item Experimental results show that {\tt CGR} can generate reasoning chains to support explainable reasoning and achieve a remarkable improvement on OpenBookQA and ARC-Challenge.
\end{itemize}

\section{Framework}
 \begin{figure*}[ht]
    \centering
\includegraphics[scale=0.5]{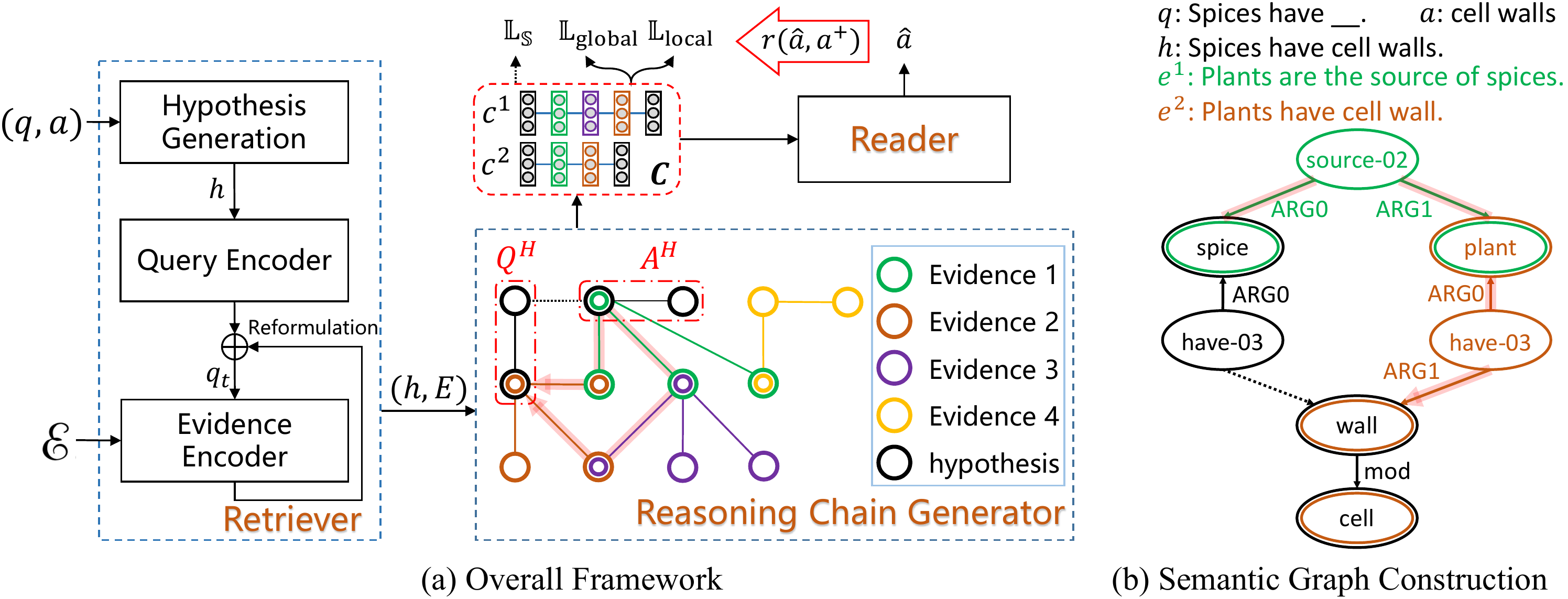} 
\caption{Overall architecture of ${\tt CGR}$. In the reasoning chain generator, the black dash lines indicate that we discard the edges between question and answer nodes. The pink arrows indicate the generated reasoning chains.}
    \label{fig:model}
\end{figure*}
\paragraph{Problem Definition} In this work, we tackle the multi-hop Science QA in the form of multi-choices, where each question $q_i$ is associated with $J$ answer choices $a_{ij}, j\in\{1,2,...,J\}$. To answer the question, one can refer to an external textual corpus $\mathcal{E}$ for relevant evidence facts. However, since the external corpus is not specifically designed for a particular multi-hop QA task, it does not necessarily contain the relevant evidence facts. Finally, based on the information at hand, our goal is to determine the correct answer.
\paragraph{Framework Overview} As shown in Figure~\ref{fig:model}-a, {\tt CGR} consists of three components: (1) A retriever iteratively retrieves a 
evidence pool $E = \{e^1,e^2,...\}$\footnote{We omit the subscript $ij$ for simplicity.} for each question-answer pair $(q,a)$ from an external textual corpus $\mathcal{E}$. (2) A reasoning chain generator first constructs a semantic graph using the fact AMRs in $E$ and then finds all complete reasoning chains $\textbf{C}=\{\textbf{c}^1,\textbf{c}^2,...\}$ on the semantic graph, where $\textbf{c}^i$ is a sequence of evidence facts. (3) A reader computes the ranking score of each answer choice, only given the facts $\hat{E} = \{\hat{e}^1,\hat{e}^2,...\}$ on these reasoning chains as the context. During training time, in addition to the standard reader loss, we propose a novel chain-aware loss that uses $\textbf{C}$ as distant supervision to train the retriever.
\subsection{Retriever}
\begin{figure}[t]
    \centering
\includegraphics[scale=0.45]{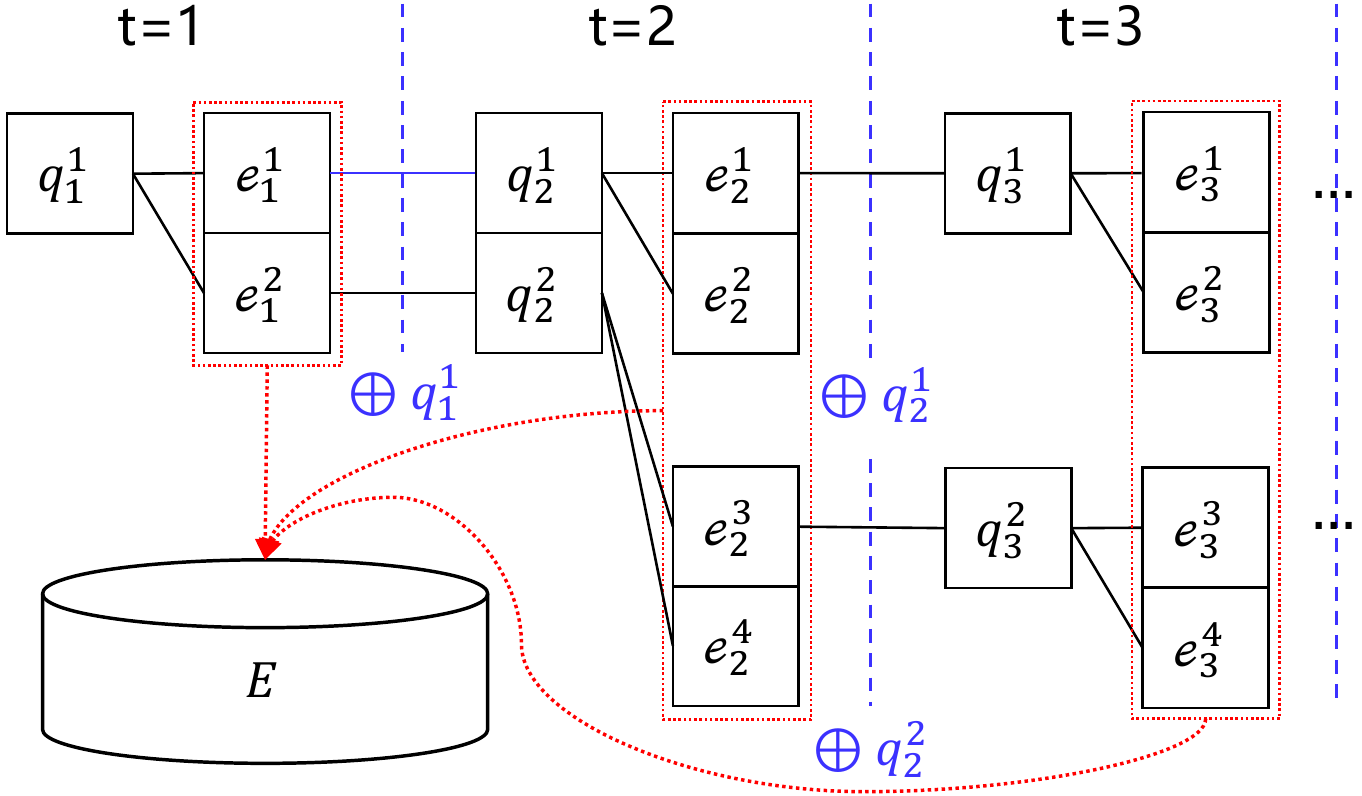} 
       \caption{Iterative retrieval process with beam size 2. }
    \label{fig:retriever}
\end{figure}
\paragraph{Hypothesis Generation} As shown in Figure~\ref{fig:model}-a, we first generate a hypothesis $h$ for each question-answer pair $(q,a)$ as the initial query for the retriever. Hypothesis generation is to convert a question-answer pair into its declarative form. Such conversion keeps all meaningful contents and maintains a good grammatical structure, which avoids noisy retrieval and allows AMR parser to generate high-quality AMR in Sec.~\ref{sec:chain}. We use the rule-based model of~\citet{demszky2018transforming} to generate the hypothesis. For unsolvable cases, we concatenate the question and the answer as the hypothesis.
\paragraph{Iterative Retrieval}
Taking into account the sequential nature of multi-hop QA, we formulate the retrieval process in an iterative fashion, where the query is updated in each iteration, conditioned on the information at hand. We adopt a similar dual-encoder of DPR~\cite{karpukhin-etal-2020-dense}. The retrieval process is accomplished by maximum inner-product search (MIPS) between the query representation and the evidence representation:
\begin{equation}
\small
    \sigma(q_t,\mathcal{e}) = E_q(q_t)^\top E_e(\mathcal{e}), \mathcal{e}\in\mathcal{E}
\end{equation}
where $E_q$ and $E_e$ are  BERT$_{\text{\tt Base}}$~\cite{devlin-etal-2019-bert} query encoder and evidence encoder respectively. We take the representation at {\tt [CLS]} as the output text representation. $q_t$ is the reformulated query in each iteration $t$. We concatenate the query at current iteration with newly retrieved evidence fact to construct the query for the next iteration:
\begin{equation}
\small
\label{equ:reformulate}
    q_t=g(q_{t-1},e_{t-1})=[q_{t-1} ; {\tt [SEP]} ; e_{t-1}]
\end{equation}
where $g(\cdot)$ is the reformulation process, $[;]$ is the concatenation. The number of iterations is $T$.  The initial query $q_1$ is the hypothesis generated above.

As shown in Figure~\ref{fig:retriever}, we introduce beam search of size $K$ to retrieve more relevant evidence facts while avoiding the search space from expanding exponentially. After the iterative retrieval, we collect evidence facts retrieved in all iterations to form a large evidence pool $E$, which is used to build the semantic graph as presented in Sec.~\ref{sec:chain}.
\subsection{Reasoning Chain Generator}
\label{sec:chain}
As shown in Figure~\ref{fig:model}, our reasoning chain generator is a non-parameterized component that generates reasoning chains using the evidence pool. It first relies on AMR to dynamically construct a semantic graph to show the relationship among the facts in the evidence pool, and then generates reasoning chains connecting both question and answer nodes. These reasoning chains serving as a byproduct of our framework can support explainability.

\paragraph{Semantic Graph Construction}
As depicted in Figure~\ref{fig:model}-b, AMR nodes are high-level abstractions of concepts conveyed in the corresponding sentence, which capture more semantic information than named entities and can be applied to different domains of sentences. We leverage on the state-of-the-art AMR parser~\cite{cai-lam-2020-amr} to generate AMRs $G=\{G^H,G^1,G^2,...\}$ for the hypothesis and its corresponding evidence pool, where $G^H$, $G^i$ are the AMR of the hypothesis and the $i^{th}$ fact in the evidence pool. The construction procedures are given as follows:

\noindent\textbf{\textit{Nodes:}} We reuse most of the concepts found in $G$ as the nodes for our semantic graph except for some over-general concepts  (e.g. {\tt (p/planet:name(n/name:op1"Earth"))}, node {\tt n/name} is an over-general concept). Fortunately, such nodes always have non-node attribute (e.g. {\tt Earth} of {\tt n/name}) that shows the specific referent. Therefore, we replace concepts with their attributes as the nodes in the semantic graph if applicable. 

\noindent\textbf{\textit{Inner-AMR Edges:}} We split the nodes of $G^H$ into question nodes $Q^H$ and answer nodes $A^H$ to represent the concepts conveyed in the question and the answer respectively. As one question is provided with $J$ answer choices, where we can generate $J$ hypothesis AMRs accordingly for each question. We take the shared nodes of $J$ hypothesis AMRs as $Q^H$, while the remaining as $A^H$:
\begin{equation}
\small
      Q^H_{ij} = \cap_{j=1}^J\{v| v \in G^H_{ij}\},
A^H_{ij} = \{v| v \in G^H_{ij}, v \notin Q^H_{ij}\}
\end{equation}
We preserve the edges in $G^H$ as edges in our semantic graph except for the edges between $Q^H$ and $A^H$ to guarantee the reasoning chains are spotted outside $G^H$. All edges in $G^i$ are preserved.

\noindent\textbf{\textit{Inter-AMR Edges:}} We add an edge between nodes of the same concept, except the root of any AMR, because these roots are mostly verbs, which may create unwanted shortcuts (e.g. {\tt have-03} in Figure~\ref{fig:model}-b). Duplicate nodes will be unified into one node, and all edges connecting those duplicate nodes are redirected to the unified node.

\paragraph{Reasoning Chain Generation}
As shown in Figure~\ref{fig:model}-a, a reasoning chain is a sequence of evidence facts that logically connect the question with the answer, while our constructed semantic graph consists of nodes in the form of AMR concepts. To tackle the mismatch, we preserve the source of each concept to show which evidence fact it comes from. During generation, we first apply depth first search on our constructed semantic graph to find all completed paths that start from one of the question nodes, pass multiple nodes outside $G^H$, and end at one of the answer nodes. We then map the nodes to their source evidence facts and thus form evidence-level reasoning chains $\textbf{C}$. We find the evidence-level reasoning chain is effective enough to capture the relationships among evidence facts and leave a more fine-grained concept-level graph in future work.

\subsection{Reader}
Existing works~\cite{chen-etal-2017-reading} typically package all the top-retrieved facts as the context for each question-answer pair, while we use the evidence facts $\hat{E}$ only on the generated reasoning chains.
\begin{equation}
\small
\label{equ:active}
    \hat{E}=  \cup_{i=1}^l\{\mathcal{e}| \mathcal{e}\in\textbf{c}^i\}
\end{equation}

We concatenate the question, the answer and all the evidence facts in $\hat{E}$ as the reader input, where {\tt [CLS]} is inserted at the beginning of the sequence, and {\tt [SEP]} are separators among the three texts. The output {\tt [CLS]} representation is fed into a classification layer, which is a linear transformation for computing the score for each question-answer pair.

\subsection{Chain-Aware Loss}
\subsubsection{Reasoning Chain Modeling}
Similar to our iterative retrieval, we model the reasoning chain in an iterative fashion. The probability of generating one reasoning chain $\textbf{c}=\{e_t\}_{t=1}^l$ is:
\begin{equation}\label{equ:modeling}
\small
    p(\textbf{c}|h)=\prod_{t=1}^{l} \frac{\text{exp}(\sigma(q_t,e_t))}{\text{exp}(\sigma(q_t,e_t))+\sum\limits_{e\in\mathcal{B}^-_t}\exp(\sigma(q_t,e))} 
\end{equation}
where $\mathcal{B}_t$ denotes all evidence facts at the $t$-th iteration and $\mathcal{B}^-_t=\mathcal{B}_t \setminus e_t$ denotes the in-batch negative examples for $e_t$, $q_t$ is the reformulated query at the $t$-th iteration from Eq. (\ref{equ:reformulate}) with $q_1=h$. 

Based on this modeling, we propose a novel chain-aware loss to train our retriever, which enables both \textbf{supervised training} with a ground-truth reasoning chain as well as \textbf{distantly supervised training} with generated reasoning chains. 

\subsubsection{Supervised Training}
Some datasets~\cite{OpenBookQA2018, khot2020qasc} annotate or partially annotate the ground-truth reasoning chain, which is a good supervision to train the retriever. 
We take the ground-truth reasoning chain as one part of our chain-aware loss if the dataset provided.  Specifically, let $\textbf{c}^+$ be an ordered ground-truth reasoning chain and $h^+$ be the generated hypothesis corresponding to the correct answer $a^+$. The supervised chain loss is defined as:
\begin{equation}
\small
  \mathbb{L}_\mathbb{S}=-\text{log}~p(\textbf{c}^+|h^+)\\
\end{equation}

\subsubsection{Distantly Supervised Training}
The ground-truth reasoning chain is quite corpus-dependent and expensive to obtain, which is not applicable in many multi-hop QA tasks~\cite{clark2018think}. 
To handle this, we leverage the generated reasoning chains from Sec.~\ref{sec:chain} as weak supervision signals to facilitate the distantly supervised training of the retriever. 
Since there are likely multiple generated reasoning chains for the same hypothesis, we elaborate two losses, namely \textbf{Local Chain Loss} and \textbf{Global Chain Loss}, to model the local chain information (in a certain reasoning chain) and the global chain information (across all the reasoning chains), respectively. 
\paragraph{Local Chain Loss (MLE)}
The local chain information is modeled in a \textit{bidirectional sequence} manner that maximizes the likelihood (MLE) between the current evidence fact with all previous evidence facts in the reasoning chain. 
Specifically, we sample up to $N$ chains $\{\hat{\textbf{c}}^i\}^N$ from the generated reasoning chains $\textbf{C}$ corresponding to $h^+$.
The forward part of MLE loss is defined by traversing each $\hat{\textbf{c}}^i$ forwardly:
\begin{equation}
\small
 \overrightarrow{\mathbb{L}_\text{mle}}=-\frac{1}{N}\sum\nolimits_{i=1}^{N}\text{log}~p(\overrightarrow{\hat{\textbf{c}}^i}|h^+)
\end{equation}

Sometimes, it may be difficult to retrieve multi-hop evidence facts from a forward direction when the question is rather confusing. Similarly, we define the backward loss $\overleftarrow{\mathbb{L}_\text{mle}}$ that traverses each $\hat{\textbf{c}}^i$ in a reversed order. 
Then the MLE loss is:
\begin{equation}
\small
    \mathbb{L}_\text{mle} = \overrightarrow{\mathbb{L}_\text{mle}} + \overleftarrow{\mathbb{L}_\text{mle}}
\end{equation}

\paragraph{Local Chain Loss (RL)} Despite the independent training of the retriever and the reader, the reader is supposed to benefit from good reasoning chains. 
To bridge such gap, we pass the information of how the reasoning chains affect the reader performance to the retriever via Reinforcement Learning (RL)~\cite{Williams-reinforcement}, such that the retriever can adapt its retrieval policy accordingly and thus allow high-quality reasoning chains to be generated. 
Specifically, given the question-answer pair as well as the corresponding reasoning chains, the \textit{action} follows the same reasoning chains sampled above.
The \textit{policy} defines the probability of bidirectionally generating those reasoning chains using Eqn.~\ref{equ:modeling} and the \textit{reward} defines the correctness of the reader prediction.
We use the following RL loss to update the parameters of the retriever via policy gradient:
\begin{equation}
\small
\begin{aligned}
     \mathbb{L}_\text{rl} =&-\frac{1}{N}\sum\nolimits_{i=1}^N [r(\hat{a},a^+)-\Bar{r}]  \cdot \\
     & [\text{log}~p(\overrightarrow{\hat{\textbf{c}}^i}|h^+) + \text{log}~p(\overleftarrow{\hat{\textbf{c}}^i}|h^+)] 
\end{aligned}
\end{equation}
where $\hat{a}$ is the predicted answer, $r(\cdot,\cdot)$ is the reward implemented as the $0/1$ indicator function. $\Bar{r}$ is the average reward in a mini-batch as the bias term.

Finally, the local chain loss is the combination of MLE and RL loss:
\begin{equation}
\small
    \mathbb{L}_\text{local}= \mathbb{L}_\text{mle}+\mathbb{L}_\text{rl}
\end{equation}
\paragraph{Global Chain Loss}
A reasoning chain usually involves indirect facts that are only related to the direct facts while share little semantic overlap with the hypothesis. Such indirect facts are unlikely to be retrieved if we fail to retrieve their corresponding direct facts. To handle this, we compute a global representation of reasoning chains by averaging the representations of all evidence facts in these chains and propose a \textit{global chain loss} to maximize the likelihood between the hypothesis and the global chain information:
\begin{equation}
\small
\begin{array}{c}
         \mathbb{L}_\text{global}=-\log\frac{\psi(\hat{E}, h^+)}{\psi(\hat{E}, h^+)+ \sum\limits_{E\in\mathcal{B}^-} \psi(E, h^+)} \\
         \psi(\hat{E}, h^+) = \exp(\frac{1}{|\hat{E}|}\sum\nolimits_{\mathcal{e} \in \hat{E}}\sigma(h^+,\mathcal{e})) 
\end{array}
\end{equation}
where we use the similar in-batch negative examples $\mathcal{B}^-=\mathcal{B}\setminus \hat{E}$ to train the global chain loss as well. $\mathcal{B}$ denotes the collection of all evidence facts in current mini-batch and $\hat{E}$ is the set of evidence facts selected by Eqn.~\ref{equ:active} for the current hypothesis $h^+$. 

\subsection{Training \& Inference}
We use the supervised chain-aware loss $\mathbb{L}_\mathbb{S}$ and the distantly supervised chain-aware loss $\mathbb{L}_\mathbb{D} = \mathbb{L}_\text{local}+\mathbb{L}_\text{global}$ to train the retriever and the standard Cross-Entropy loss $\mathbb{L}_\text{reader}$ between the reader prediction and the correct answer choice to train the reader. The final training objective is:
\begin{equation}
\small
    \mathbb{L}=\mathbb{L}_\text{reader}+\mathbb{L}_\mathbb{S}+\mathbb{L}_\mathbb{D}
\end{equation}

During inference, each question-answer pair follows the same pipeline of our retriever, reasoning chain generator, and reader to get its ranking score. The top-ranked choice is chosen as the output.

\section{Experimental Setup}
\paragraph{Datasets:} We evaluate the effectiveness of our {\tt CGR} framework on two multi-hop science QA datasets: OpenBookQA~\cite{OpenBookQA2018} and ARC-Challenge~\cite{clark2018think}, where the ground-truth reasoning chain is either partially provided or not provided. OpenBookQA and ARC-Challenge provide their corresponding leaderboards with train, develop and test sets publicly available. Following~\citet{AristoRoBERTAv7}, we combine the training set of OpenBookQA (4957), ARC-Easy (2251), ARC-Challenge (1119) and \textit{RegLivEnv} (665) as the final training set of ARC-Challenge task. The data splits is shown in Table~\ref{tab:dataset}. OpenBookQA annotates one evidence fact on the ground-truth reasoning chain for each question-answer pair, which is used in $\mathbb{L}_\mathbb{S}$. ARC-Challenge does not provide the ground-truth reasoning chain, where $\mathbb{L}_\mathbb{S}$ is eliminated from the final training objective. The textual corpus is ARC Corpus~\cite{clark2018think} for both two tasks, consisting of 14M science facts. Moreover, OpenBookQA provides an accompanying open-book with 1326 science facts, which are highly related to the questions in this dataset. Therefore, we performed an extra retrieval in the open-book in the first iteration.
\begin{table}[t]
    \centering
    \begin{tabular}{cccc}
    \toprule
         & Train & Dev & Test\\ \hline
         OpenBookQA& 4957 &500 &500\\
         ARC-Challenge& 8992 &299 &1172\\ \bottomrule
    \end{tabular}
    \caption{Number of instances in each dataset.}
    \label{tab:dataset}
\end{table}

\paragraph{Implementation:} We use AristoRoBERTa~\cite{AristoRoBERTAv7} as our reader and Sentence-BERT~\cite{reimers-2019-sentence-bert} as our retriever. As indicated by~\citet{NEURIPS2020_6b493230}, training the evidence encoder $E_e$ that requires periodic updates of the evidence index is costly and does little improvement to the performance. Therefore, we fix the evidence encoder and cache all evidence representations offline for efficiency purposes. We set the beam size $K$ to 10 and the total iteration step $T$ to 2, resulting in an average size of evidence pool to be 78 and 53 for OpenBookQA and ARC-Challenge respectively. We then select facts for the reader with maximum evidence size of 15 and 20 respectively. The number of sampled chains $N$ is set to 1.~\footnote{Our code is available at: \url{https://github.com/wwxu21/CGR}}
More details and analysis can be found in Appendix~\ref{app:impl} and ~\ref{sec:KT}.
\paragraph{Comparison Methods:} For fair comparison, we compare {\tt CGR} with recently published methods that use similar power of pretrained models, including five textual knowledge based methods: AristoRoBERTa~\cite{AristoRoBERTAv7}, KF-SIR~\cite{banerjee2020knowledge}, FreeLB~\cite{Zhu2020FreeLB}, DPR~\cite{karpukhin-etal-2020-dense}, AMR-SG~\cite{xu-etal-2021-dynamic} and another two methods leveraging on an additional knowledge graph~\cite{speer2017conceptnet}: PG~\cite{wang-etal-2020-connecting:}, and MHGRN~\cite{feng-etal-2020-scalable}.
\section{Results}
\subsection{QA Performance}
\begin{table}[t]
\small
    \centering
    \begin{tabular}{lp{13mm}<{\centering}p{12mm}<{\centering}c}
    \toprule
      \normalsize{Methods}&  Additional KG & Output Chains & Test Acc.  \\\midrule
        PG& \checkmark& \checkmark &81.8\\
        AMR-SG & $\times$& $\times$ &81.6\\
        DPR     & $\times$& $\times$&80.8 \\
         MHGRN&\checkmark& \checkmark  &80.6\\
         KF-SIR & $\times$ & $\times$&80.0\\ \midrule
          AristoRoBERTa& $\times$& $\times$ & 77.8\\ 
         + {\tt CGR} & $\times$& \checkmark & \textbf{82.4}\\ \bottomrule
    \end{tabular}
    \caption{Test accuracy on OpenBookQA. Methods that use additional KG or can output reasoning chains are ticked respectively.}
    \label{tab:leadobqa}
\end{table}
\begin{table}[t]
    \centering
    \small
    \begin{tabular}{lp{13mm}<{\centering}p{12mm}<{\centering}c}
    \toprule
      Methods & Additional KG & Output Chains  &Test Acc.  \\
 \midrule
 AMR-SG & $\times$& $\times$& 68.94 \\
        FreeLB & $\times$& $\times$& 67.75\\
        arcRoberta $\spadesuit$ & $\times$ & $\times$& 67.15\\ 
        xlnet+Roberta $\spadesuit$ & $\times$& $\times$& 67.06\\ \midrule
        AristoRoBERTa & $\times$& $\times$& 66.47\\ 
        + {\tt CGR}& $\times$ & \checkmark& \textbf{69.20} \\\bottomrule
    \end{tabular}
    \caption{Test accuracy on ARC-Challenge. $\spadesuit$ are unpublished methods.}
    \label{tab:arc}
\end{table}
\paragraph{OpenBookQA:} Table~\ref{tab:leadobqa} shows the comparison results of OpenBookQA. Our {\tt CGR} significantly improves over the baseline AristoRoBERTa with 4.6 accuracy score. Meanwhile, {\tt CGR} can also provide the reasoning chains as an additional output that reflect the step-by-step reasoning process to infer the answer, making the QA process explainable. 
When compared to recently published methods, we find that {\tt CGR} can also surpass methods leveraging on additional KG. It suggests that textual knowledge resource is still under-investigated, where the gap between the query and indirect fact is one of the issues that restricts the retriever performance for multi-hop QA. UnifiedQA~\cite{khashabi-etal-2020-unifiedqa} and T5 3B~\cite{raffel2020exploring} are two extremely large models (with 30x more parameters than other models), which are not fair for comparison.

\paragraph{ARC-Challenge:}
We implement {\tt CGR} on another task: ARC-Challenge, where the ground-truth reasoning chain is not available. As shown in Table~\ref{tab:arc}, our {\tt CGR} significantly improves the baseline AristoRoBERTa with 2.73 accuracy score, which demonstrates the effectiveness of {\tt CGR} in generating and modeling the reasoning chain in a more general manner. Notably, {\tt CGR} achieves a new state-of-the-art performance in this challenging task in a computationally practicable setting.

\begin{table}[t]
    \centering
    \small
    \begin{tabular}{lcc}
    \toprule
    Methods & ~~~~~~~~~~~~w/o $\mathbb{L}_\mathbb{S}$ ~~~~~~~~~~~~& w/ $\mathbb{L}_\mathbb{S}$ \\ \midrule
        {\tt CGR} &79.70$\pm$0.33 &\textbf{81.70$\pm$0.49} \\
        - $\mathbb{L}_\text{global}$&78.60$\pm$0.28 &80.95$\pm$0.30\\
        - $\mathbb{L}_\text{local}$ {\small (MLE)} &79.10$\pm$0.30 &80.55$\pm$0.52 \\
        - $\mathbb{L}_\text{local}$ {\small (RL)}  &78.70$\pm$0.36 &80.80$\pm$0.22 \\
        - $\mathbb{L}_\text{local}$ {\small (Both)}&78.40$\pm$0.37 &81.05$\pm$0.30\\
        - $\mathbb{L}_\mathbb{D}$                     &78.20$\pm$0.37    &80.55$\pm$0.38\\
    \bottomrule
    \end{tabular}
    \caption{Ablation study on the composition of the training objectives on OpenBookQA. Accuracy (mean$\pm$ standard deviation) are computed over 4 runs.}
    \label{tab:abl-retriever}
\end{table}
\paragraph{Ablation Study:} Table~\ref{tab:abl-retriever} shows our ablation study on the composition of the training objectives both in the presence or absence of the ground-truth reasoning chain on OpenBookQA. 
We can observe the same performance trend under two scenarios. First, we observe a degradation of QA performance when removing $\mathbb{L}_\text{global}$. 
As $\mathbb{L}_\text{global}$ provides a rough idea of the reasoning chain, it reduces the difficulty to retrieve indirect facts. Moreover, $\mathbb{L}_\text{global}$ is still important even the ground-truth evidence chain is present because it improves the generalization of our framework to retrieve other reasoning chains that can answer the question rather than overfitting to the ground-truth reasoning chain.
Second, discarding $\mathbb{L}_\text{local}$ also casts a negative impact on the QA performance. $\mathbb{L}_\text{local}$, on the other hand, is a fine-grained modeling of evidence-evidence interactions. It is effective to distinguish the correct answer because the incorrect ones would get a relatively low probability to form the reasoning chain.
Third, discarding both global and local chain losses results in more severe performance degradation, which demonstrates the necessity of our modeling for the reasoning chains both globally and locally.
\subsection{Iterative Retriever Performance}
\begin{table}[t]
    \centering
    \small
    \begin{tabular}{lcccc}
    \toprule
    IR Methods            & Acc. & Dir. & Ind.& Com. \\ \midrule
       TF-IDF             & 52.4 & 3.72 & 0.38 & 0.46 \\
       Dense Vector       & 54.6 & 4.68 & 0.40 & 0.68\\
       Iterative Retrieval& \textbf{63.6} & 5.00 & 0.48 & 0.70\\
       Chain Generator    & 60.2 & \textbf{5.24} & \textbf{0.96} & \textbf{0.74}\\
        \bottomrule
    \end{tabular}
    \caption{Automatic and Human Evaluations of the IR performance on OpenBookQA. }
    \label{tab:retriever}
\end{table}

As mentioned, one of our major contributions is to form reasoning chains that capture both direct and indirect evidence facts. To evaluate the quality of the retrieved facts, we conduct both automatic and human evaluations. As OpenBookQA provides one ground-truth evidence fact, we use the retrieval accuracy (Acc.) as our automatic evaluation. For human evaluation, we evaluate the quality from three aspects: (1) \textit{Directly-Related (Dir.):} The evidence is a direct fact and is useful to answer the question. (2) \textit{Indirectly-Related (Ind.):} The evidence is an indirect fact and is useful to answer the question. (3) \textit{Completeness (Com.):} All evidence facts in $\hat{E}$ can together form a real reasoning chain that completely fills the knowledge gap between the question and the answer. We randomly sample 50 questions and evaluate the quality of evidence facts corresponding to the correct answer choice, where each fact contributes 1 score if it meets the requirement of \textit{Dir.} and \textit{Ind.} (ranging from 0 to 15), and all evidence facts in $\hat{E}$ contribute 1 score if they together meet \textit{Com.} (ranging from 0 to 1).

As shown in Table~\ref{tab:retriever}, we conduct evaluations on four Information Retrieval (IR) methods. Among those IR methods, Dense Vector retriever is effective in finding direct facts than word-match based retriever (TF-IDF) but faces the same limitation in finding indirect facts. Iterative Retrieval can remedy this to some extent, but it is a relatively loose restriction, where the retrieved facts can be biased to some particular facts. Surprisingly, our Reasoning Chain Generator significantly improves the recall of retrieving indirect facts with \textit{Ind.} almost doubled. Though Reasoning Chain Generator may hurt \textit{Acc.}, the improvements on both \textit{Dir.} and \textit{Ind.} show that it can still find alternative facts from the textual corpus to form the reasoning chain.
\section{Discussions on Explainability}
\begin{figure}[t]
    \centering
\includegraphics[scale=0.6]{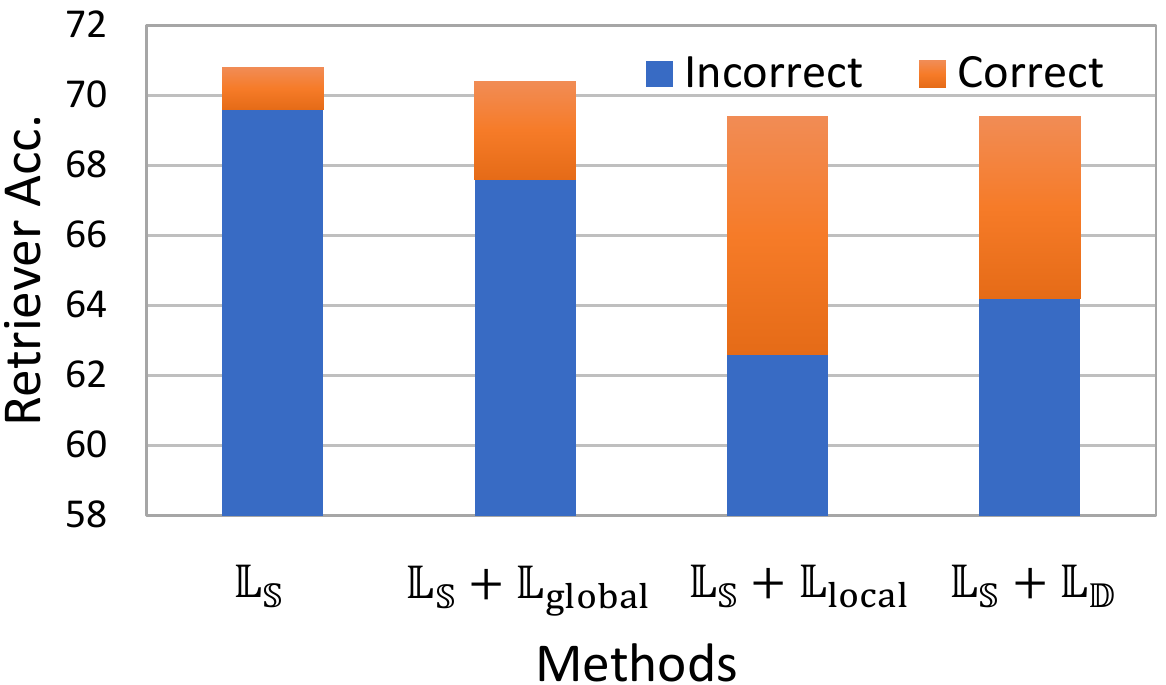} 
\caption{Retriever accuracy on OpenBookQA. The ground-truth evidence in the context of (1) the correct answer choice; (2) any three incorrect answer choices.}
    \label{fig:retr_dif}
\end{figure}
\subsection{Effect of Chain Modeling}
We plot the accuracy of the ground-truth evidence fact retrieved either by the hypothesis corresponding to the correct answer and the hypotheses corresponding to incorrect answers in Figure~\ref{fig:retr_dif}. Firstly, though the main purpose of $\mathbb{L}_\text{global}$ is to improve the generalization ability of our framework, it can also slightly reduce the retrieval accuracy for incorrect answers with little hurt to the correct answer. 
Secondly, $\mathbb{L}_\text{local}$ significantly reduces the retrieval accuracy for incorrect answers.
It makes the incorrect answers less deceptive and thus makes it much easier for the reader to distinguish the correct answer, which is concise with the results in Table~\ref{tab:abl-retriever}. 
Thirdly, the combination of the two chain-aware losses may affect the retriever performance marginally, but in terms of overall QA performance, they obviously bring benefits.
 \begin{figure}[t]
    \centering
\includegraphics[scale=0.53]{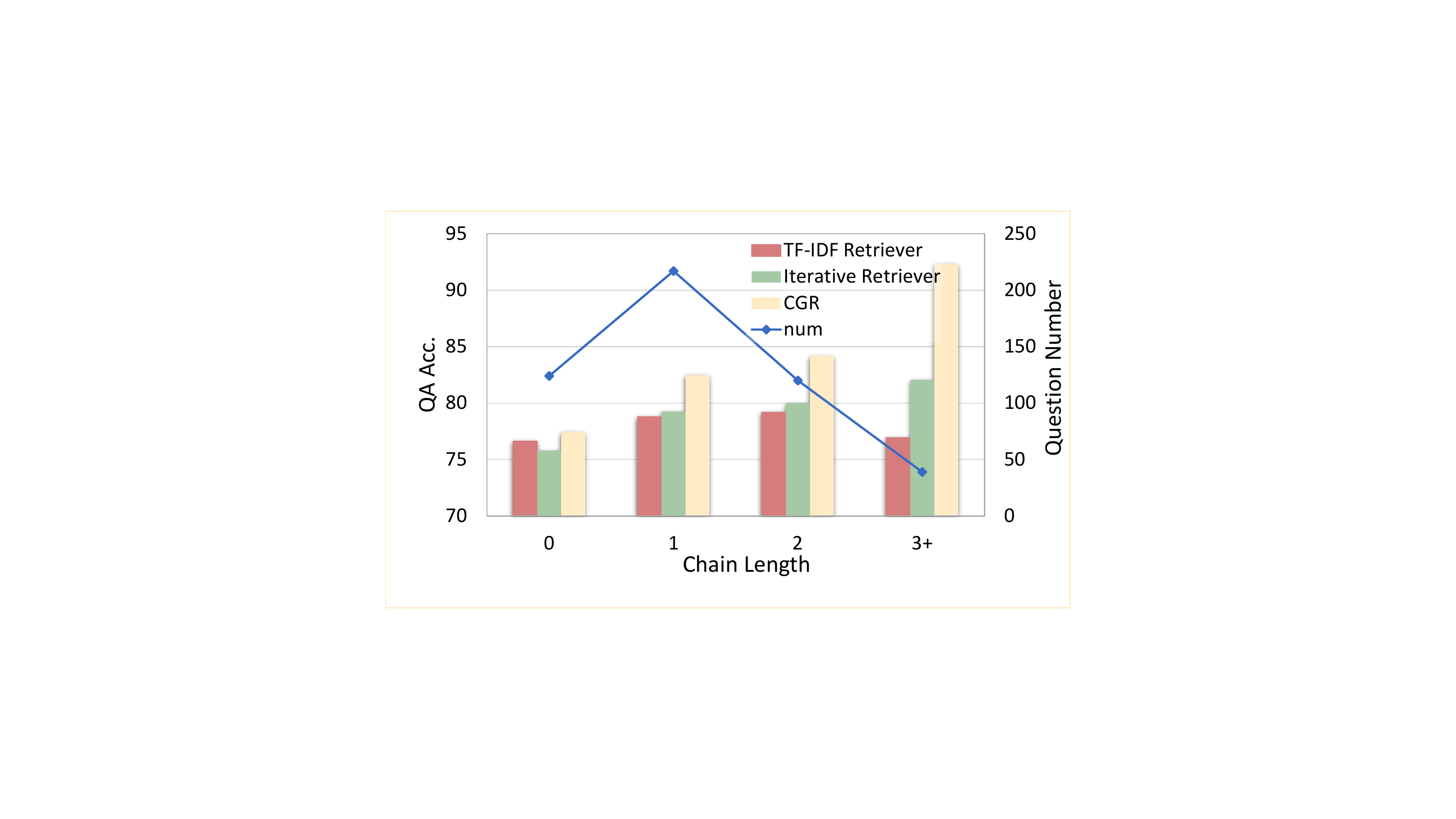} 
\caption{QA performance in terms of reasoning chain length on OpenBookQA. TF-IDF and iterative retriever only use retriever-reader architecture.}
    \label{fig:chain}
\end{figure}
\subsection{Scalability on Chain length}
We plot the QA performance in terms of different chain lengths on OpenBookQA in Figure~\ref{fig:chain}. As we may generate multiple reasoning chains with different lengths for each question-answer pair, we define the length of the reasoning chain for each question as the most frequent length with respect to the correct answer. Length of 0 indicates that we cannot generate any reasoning chain for these questions, because (1) some decisive evidence facts are missing in our textual corpus, and (2) the performance of the AMR parser also limits the upper bound of our reasoning chain generator. Meanwhile, such cases are also difficult for TF-IDF and iterative retriever, which should be highly considered in the future. Apart from this, we can observe a comprehensive improvement of {\tt CGR} in handling multi-hop QA. Such improvement is much more obvious when the chain length becomes larger, which is commonly hard for reasoning. This demonstrates the effectiveness of our explicit modeling for the reasoning chain, especially when the question is more complex.
\subsection{Case Study on Reasoning}
\begin{table}[]
    \centering
    \small
        \begin{tabular}{|p{73mm}|}
    \hline
         \textbf{Question:} \textit{A person wants to be able to have more natural power in their home. They choose to cease using a traditional electric company to source this electricity, and so decide to install} (A) sun grafts (B) sunlight shields \textbf{(C) panels collecting sunlight} (D) solar bees \\ \hline
         \textbf{Useful facts retrieved by iterative retrieval:} \\
         	{[1]} A solar panel converts sunlight into electricity. \\ \hline
         \textbf{Additional facts from reasoning chain generator:} \\
        {[2]} Solar energy is a renewable resource.\\
{[3]} Such renewable resources are called, natural resources. \\ \hline
         \textbf{Reasoning Chain}: \\
         Question $\xrightarrow{\tt natural-03}$ {[3]} $\xrightarrow{\tt renew-01}$ {[2]} $\xrightarrow{\tt solar}$ {[1]} $\xrightarrow{\tt panel}$ (C)\\
         \hline
    \end{tabular}
\caption{Case study on OpenBookQA.}
\label{case_study}
\end{table}
We show how our generated reasoning chain can support explainable reasoning in Table~\ref{case_study}. Iterative retrieval can retrieve the first evidence fact as it shares critical semantic information {\tt panel} with the choice {\tt C}. However, it fails to retrieve the second and the third evidence fact because (1) the second one is an indirect fact sharing little semantic overlap with either the question or the answer choice, and (2) the third one though serves as a direct fact, it shows a relatively low similarity with the question due to the massive information conveyed in the question. On the other hand, our reasoning chain is able to discover evidence facts that fail to be retrieved by an iterative retriever and form a reasoning chain with AMR concepts as anchors.
As the incorrect answers are not likely to form reasoning chains, the evidence facts on the reasoning chains are highly discriminative and can effectively support the reader to select the correct answer. More cases can be found in Appendix~\ref{app:more_case}.
\section{Related Work}
\paragraph{Multi-hop QA:} Multi-hop QA is a challenging task as it requires gathering multiple evidence facts, especially indirect facts, to form a reasoning chain. Early attempts mostly rely on iterative retrieval. For example,\citet{yadav-etal-2019-quick} extract evidence facts in consideration of their relevance, overlap and coverage. \citet{banerjee-etal-2019-careful, yadav-etal-2020-unsupervised} reformulate their queries with unused words in the last iteration. However, these methods may retrieve irrelevant facts as the query grow biased to unimportant words. As some recent QA datasets annotate the ground-truth reasoning chain~\cite{yang-etal-2018-hotpotqa, OpenBookQA2018}, they enable training supervised classifier to identify the correct evidence facts~\cite{nie-etal-2019-revealing, tu2020select}. It is a good step to control the quality of reasoning chains, but still remains an issue when the ground-truth reasoning chain is not available~\cite{clark2018think}. Other works explore the effectiveness of KG by either automatically constructing the graph using named entity or semantic role labeling~\cite{qiu-etal-2019-dynamically, bosselut-etal-2019-comet, fang-etal-2020-hierarchical, chen2019multi} or resorting to existing KG~\cite{saxena-etal-2020-improving,zhang2020gmh,yasunaga2021qa}.  Despite the high precision of those KGs, they are known to suffer from sparsity in existing KG~\cite{zhao2020complex}, where complex reasoning chains are unlikely to be covered by the closed-form relations in KG~\cite{lei2020interactive}.

\paragraph{Dense-Vector Retriever:} In contrast to term-based retriever implemented with TF-IDF or BM25~\cite{chen-etal-2017-reading,wang2018r}, dense-vector  retriever has received increasing attention as it captures the semantic matching beyond simple word overlap and can be trained along with the reader~\cite{zhu2021retrieving}. It has been reported to outperform term-based methods in many open-domain QA tasks~\cite{das2018multistep,karpukhin-etal-2020-dense,min2021neurips}, including those on multi-hop QA~\cite{Asai2020Learning,xiong2021answering}.

\paragraph{AMR:} AMR has been successfully coupled with many natural language processing tasks in explicit reasoning, such as summarization~\cite{liao-etal-2018-abstract}, event detection~\cite{li-etal-2015-improving-event}, machine translation~\cite{song-etal-2019-semantic}, and symbolic QA~\cite{kapanipathi2020question}. Comparing to named entity~\cite{shao-etal-2020-graph}, we use AMR as our graph annotation because it is not specifically designed for a particular domain and can be adapted to a wide range of textual data.
\section{Conclusion}
We propose a novel Chain Guided Retriever-reader framework for multi-hop QA. Our modeling for the reasoning chains is effective to find both direct and indirect facts and is less likely to introduce noise. Moreover, our framework is corpus-independent and is capable of handling the setting without any ground-truth annotations. Further analysis and discussions also elucidate some of the inner workings of our framework while maintaining the explainability at the same time.

\bibliography{anthology,custom}
\bibliographystyle{acl_natbib}
\appendix
\clearpage
\section{Appendix}
\subsection{Implementation}
\label{app:impl}
For OpenBookQA, we implement our {\tt CGR} framework only in the last fine-tuning process as indicated in \citet{AristoRoBERTAv7}, where only OpenBookQA dataset is used. The initial learning rate is 9e-6, the batch size is 4 and the max sequence length is 256. For ARC-Challenge, we implement our {\tt CGR} framework on the combination of all above training sets. The initial learning rate, the batch size and the max sequence length are 1e-5, 6, and 384 respectively. We use grid search to find optimal hyper-parameters, where the learning rate is chosen from \{5e-6,8e-6,9e-6,1e-5,1.1e-5, 1.2e-5,1.5e-5\}, the batch size is chosen from \{4,6,8,12,16\}, beam size $K$ is chosen from \{5,10,15,20\} and iteration step $T$ is chosen from \{1,2,3\}. 
 
We introduce 110M parameters of our retriever in addition to 355M of our reader. We run all experiments on one TITAN RTX card, which takes about 2 hour and 8 hours to complete the training of OpenBookQA and ARC-Challenge respectively.

\subsection{Effect of Beam size and Iteration Step}
\label{sec:KT} 
\begin{figure}[h]
    \centering
\includegraphics[scale=0.6]{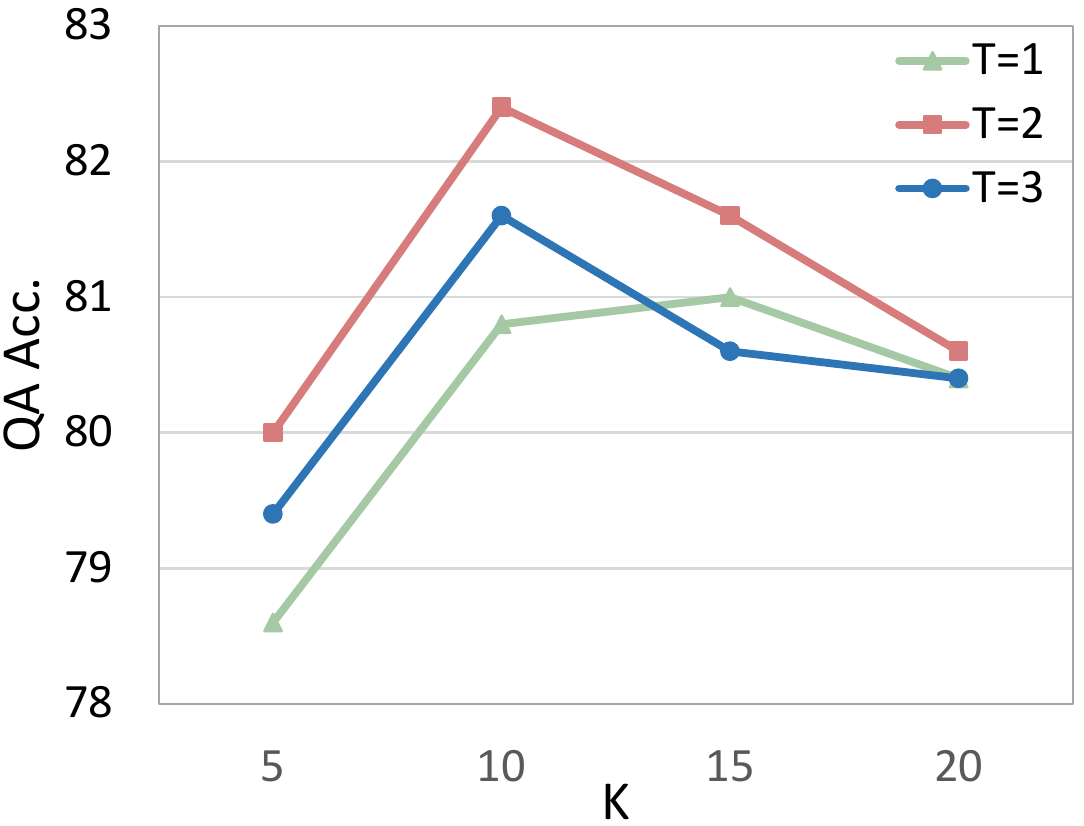} 
\caption{Effect of hyper-parameter Beam size ($K$) and Iteration Step ($T$) on OpenBookQA.}
    \label{fig:KT}
\end{figure}

We vary two hyper-parameters $K$ and $T$ to show their effects on OpenBookQA. As depicted in Figure~\ref{fig:KT}, the model with $T=1$ has a relatively lower performance than the other two models because it suffers from a low recall of relevant evidence facts, which also explains why it benefits more from a larger $K$. Moreover, model with $T=2$ performs better than model with $T=3$. It indicates most of the questions can be solved with a reasoning chain of length 2, which is consistent with the construction of this dataset. In addition, models with $T>1$ reaches the top at $K=10$. This might be due to more noisy retrievals in a larger evidence pool.

\subsection{Case Study}
\label{app:more_case}
More case studies can be found in Table~\ref{tab:more_case}.
\begin{table*}[]
    \centering
\subtable[Case Study 1]{
        \begin{tabular}{|p{40mm}|p{103mm}|}
    \hline
    \textbf{Question:} & \textit{Which requires energy to move?} \textbf{(A) weasel} (B)willow (C) mango (D) poison ivy \\ \hline
         \textbf{Related Evidence Facts:} & {[1]} An animal requires energy to move.
         
 {[2]} Predator is a animal.
 
 {[3]} A weasels food chain is a predator.\\ \hline
         \textbf{Reasoning Chain}: & Question $\xrightarrow{\tt energy}$ {[1]} $\xrightarrow{\tt animal}$ {[2]} $\xrightarrow{\tt predator}$ {[3]} $\xrightarrow{\tt weasel}$ (A) \\
         \hline
    \end{tabular}
         
\label{3table}}
\subtable[Case Study 2]{
        \begin{tabular}{|p{40mm}|p{103mm}|}
    \hline
         \textbf{Question:} & \textit{A positive effect of burning biofuel is} (A) shortage of crops for the food supply (B) an increase in air pollution \textbf{(C) powering the lights in a home} (D) deforestation in the amazon to make room for crops
  \\ \hline
         \textbf{Related Evidence Facts:} & {[1]} Biofuel is used to produce electricity by burning.
         
 {[2]} Some light bulbs convert electricity into light and heat energy. \\ \hline
         \textbf{Reasoning Chain}: & Question $\xrightarrow{\tt biofuel}$ {[1]} $\xrightarrow{\tt electricity}$ {[2]} $\xrightarrow{\tt light}$ (C) \\
         \hline
    \end{tabular}
\label{3table}}
\subtable[Case Study 3]{
    \begin{tabular}{|p{40mm}|p{103mm}|}
    \hline
         \textbf{Question:} & \textit{An example of conservation is avoiding the use of} \textbf{(A) gasoline} (B) air (C) snow (D) clothes \\ \hline
         \textbf{Related Evidence Facts:} & {[1]} An example of conservation is not using fossil fuel.
         
 {[2]} Gasoline is a fuel mixture. \\ \hline
         \textbf{Reasoning Chain}: & Question $\xrightarrow{\tt conserve-01}$ {[1]} $\xrightarrow{\tt fuel}$ {[2]} $\xrightarrow{\tt gasoline}$ (A) \\
         \hline
    \end{tabular}
\label{firsttable}}
\subtable[Case Study 4]{
    \begin{tabular}{|p{40mm}|p{103mm}|}
    \hline
         \textbf{Question:} & \textit{They studied the soil by using} (A) plants (B) a telescope (C) roots \textbf{(D) a microscope}
  \\ \hline
         \textbf{Related Evidence Facts:} & {[1]} Studying a soil sample means studying the small microorganisms in that soil.
         
 {[2]}  Magnifying makes seeing microorganisms easier through using a microscope. \\ \hline
         \textbf{Reasoning Chain}: & Question $\xrightarrow{\tt soil}$ {[1]} $\xrightarrow{\tt microorganism}$ {[2]} $\xrightarrow{\tt microscope}$ (D) \\
         \hline
    \end{tabular}
\label{firsttable}}

\subtable[Case Study 5]{
    \begin{tabular}{|p{40mm}|p{103mm}|}
    \hline
        \textbf{Question:} & \textit{Birds carrying away fruit helps the tree} (A) grow (B) fertilize \textbf{(C) reproduce} (D) conquer  \\ \hline
         \textbf{Related Evidence Facts:} & {[1]} Birds are a vehicle for spreading the seeds of a plant.
         
 {[2]} Ex2: plants reproduce with seeds. \\ \hline
         \textbf{Reasoning Chain}: & Question $\xrightarrow{\tt bird}$ {[1]} $\xrightarrow{\tt plant}$ {[2]} $\xrightarrow{\tt reproduce-01}$ (C) \\
         \hline
    \end{tabular}
\label{firsttable}}
\subtable[Case Study 6]{
    \begin{tabular}{|p{40mm}|p{103mm}|}
    \hline
        \textbf{Question:} & \textit{The salamander could eat a large amounts of what?} (A) fettuccine (B) waxy leaves from certain plants (C) dead carcass meat from livestock \textbf{(D) six legged winged organisms}  \\ \hline
         \textbf{Related Evidence Facts:} & {[1]} A salamander eats insects.
         
 {[2]} Insects have three parts to their bodies, wings, two feelers, and six legs. \\ \hline
         \textbf{Reasoning Chain}: & Question $\xrightarrow{\tt salamander}$ {[1]} $\xrightarrow{\tt insect}$ {[2]} $\xrightarrow{\tt wing}$ (D) \\
         \hline
    \end{tabular}
\label{firsttable}}
    \caption{More case studies in addition to Table~\ref{case_study}}
    \label{tab:more_case}
\end{table*}
\end{document}